\documentclass[]{style}
\usepackage[toc,page,header]{appendix}
\usepackage{wrapfig}

\usepackage{booktabs}
\usepackage{multirow}
\usepackage[table]{xcolor}
\usepackage{colortbl}
\usepackage{array}
\usepackage{makecell}

\definecolor{headerblue}{RGB}{220,230,242}
\definecolor{groupgray}{RGB}{242,244,247}
\definecolor{semigreen}{RGB}{232,245,233}
\definecolor{bestred}{RGB}{255,235,238}
\definecolor{lightblue}{RGB}{245,249,255}

\usepackage{booktabs}
\usepackage{multirow}
\usepackage[table]{xcolor}
\usepackage{colortbl}
\usepackage{pifont}

\definecolor{headerblue}{RGB}{220,230,242}
\definecolor{groupgray}{RGB}{242,244,247}
\definecolor{semigreen}{RGB}{232,245,233}
\definecolor{bestred}{RGB}{255,235,238}
\definecolor{secondblue}{RGB}{232,240,254}
\definecolor{lightblue}{RGB}{245,249,255}

\newcommand{\best}[1]{\cellcolor{bestred}\textbf{#1}}
\newcommand{\second}[1]{\cellcolor{secondblue}{#1}}
\newcommand{\semirow}{\rowcolor{semigreen}}
\newcommand{\grouprow}{\rowcolor{groupgray}}
\providecommand{\cmark}{\ding{51}}
\providecommand{\xmark}{\ding{55}}
\usepackage{natbib}

\usepackage{CJKutf8}

\usepackage{xargs}

\usepackage{todonotes}

\usepackage{multirow}

\usepackage{cleveref}

\usepackage{amsmath}
\usepackage{amssymb}
\usepackage{mathtools}
\usepackage{amsthm}
\usepackage{dsfont}

\usepackage{svg}

\usepackage{mathrsfs}
\usepackage{adjustbox}
\usepackage{multicol}
\usepackage{tcolorbox}
\usepackage{changepage}
\usepackage{enumitem}
\usepackage{graphicx}
\usepackage{xcolor}
\usepackage{float}
\usepackage{threeparttable}
\usepackage{subcaption}
\usepackage{algorithm}
\usepackage{algpseudocode}
\usepackage{wrapfig}
\usepackage[table]{xcolor}
\usepackage{colortbl}
\usepackage{tabularx}
\usepackage{makecell}
\usepackage[dvipsnames]{xcolor}

\newcolumntype{g}{>{\columncolor{gray!10}}c}

\theoremstyle{plain}

\theoremstyle{definition}

\theoremstyle{remark}

\definecolor{catgray}{gray}{0.9}
\definecolor{skyblue}{rgb}{0.53,0.81,0.92}
\colorlet{skyblue!30}{skyblue!30!white}
\definecolor{customblue}{RGB}{70,130,180}

\providecommand{\cmark}{\textcolor{green!60!black}{\checkmark}}
\providecommand{\xmark}{\textcolor{red!70!black}{\ding{55}}}

\renewcommand{\emph}[1]{\textit{#1}}

\usepackage{minitoc}
\usepackage{url}

\PassOptionsToPackage{table,xcdraw}{xcolor}
\usepackage{titletoc}
\usepackage{placeins}
\usepackage{pifont}

\definecolor{RowBlue}{HTML}{E9F2FB}
\definecolor{RowRed}{HTML}{F9EAEA}
\definecolor{Top1}{HTML}{50DB4B}
\definecolor{Top2}{HTML}{A5FFA2}
\definecolor{Top3}{HTML}{D9FFD9}
\definecolor{Sub1}{HTML}{EAB8B8}
\definecolor{Sub2}{HTML}{E4E4E4}

\title{Semi-Supervised Vision-Language-Action Model}

\author[1,3\dagger]{Hongyang He}
\author[2]{Jiuming Liu}
\author[1]{Victor Sanchez}

\affiliation[1]{University of Warwick}
\affiliation[2]{University of Cambridge}
\affiliation[3]{Manifolda.Ai}

\contribution[\dagger]{Corresponding author}

\abstract{
Vision-Language-Action (VLA) models enable robots to predict actions directly from visual observations and language instructions, but adapting them to new environments still depends on costly action-labeled demonstrations. 
To reduce this dependence, we study semi-supervised VLA adaptation under limited supervision signals, where only a small portion of trajectories contain robot actions and the remaining trajectories provide action-unlabeled vision-language observations. 
Unlike standard semi-supervised learning, the missing supervision is an embodied action signal that must be visually grounded, language-consistent, physically feasible, and temporally stable. 
To address this problem, we propose SemiVLA, a self-distilled teacher-student framework that learns from reliable pseudo-actions on unlabeled trajectories. 
SemiVLA introduces a VLA-specific reliability controller to assess vision-language alignment, action feasibility, and temporal transition consistency, and further updates the teacher through a Bottleneck-Projected Alignment Update to avoid noisy feedback contamination. 
With OpenVLA as the backbone, SemiVLA consistently improves multiple PEFT strategies across LIBERO and CALVIN. 
Under 10\% labeled trajectories, SemiVLA with Selective LoRA achieves 89.0\% average success on LIBERO, outperforming supervised LoRA by 8.0 points without extra inference cost.
}

\begin{document}
\maketitle

\section{Introduction}
\label{sec:intro}

\begin{figure*}[t]
    \centering
    \includegraphics[width=\linewidth]{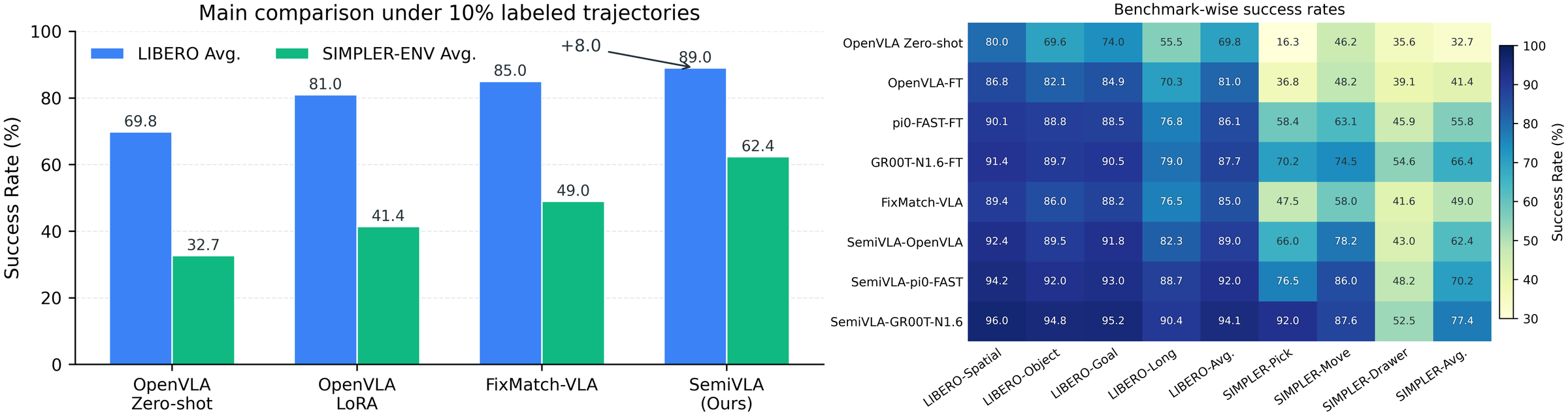}
    \caption{
    \textbf{Main benchmark visualization under the proposed semi-supervised VLA setting.}
    Left: comparison of average performance on LIBERO and SIMPLER-ENV under 10\% labeled trajectories.
    SemiVLA consistently improves over zero-shot OpenVLA, supervised LoRA adaptation, and confidence-based semi-supervised baselines.
    Right: benchmark-wise heatmap across LIBERO and SIMPLER-ENV task suites.
    SemiVLA achieves stronger performance across both short-horizon and long-horizon manipulation tasks, showing that reliable pseudo-action learning improves adaptation beyond a single benchmark or task type.
    }
    \label{fig:main_results_visual}
\end{figure*}

\begin{figure}[t]
    \centering
    \includegraphics[width=\linewidth]{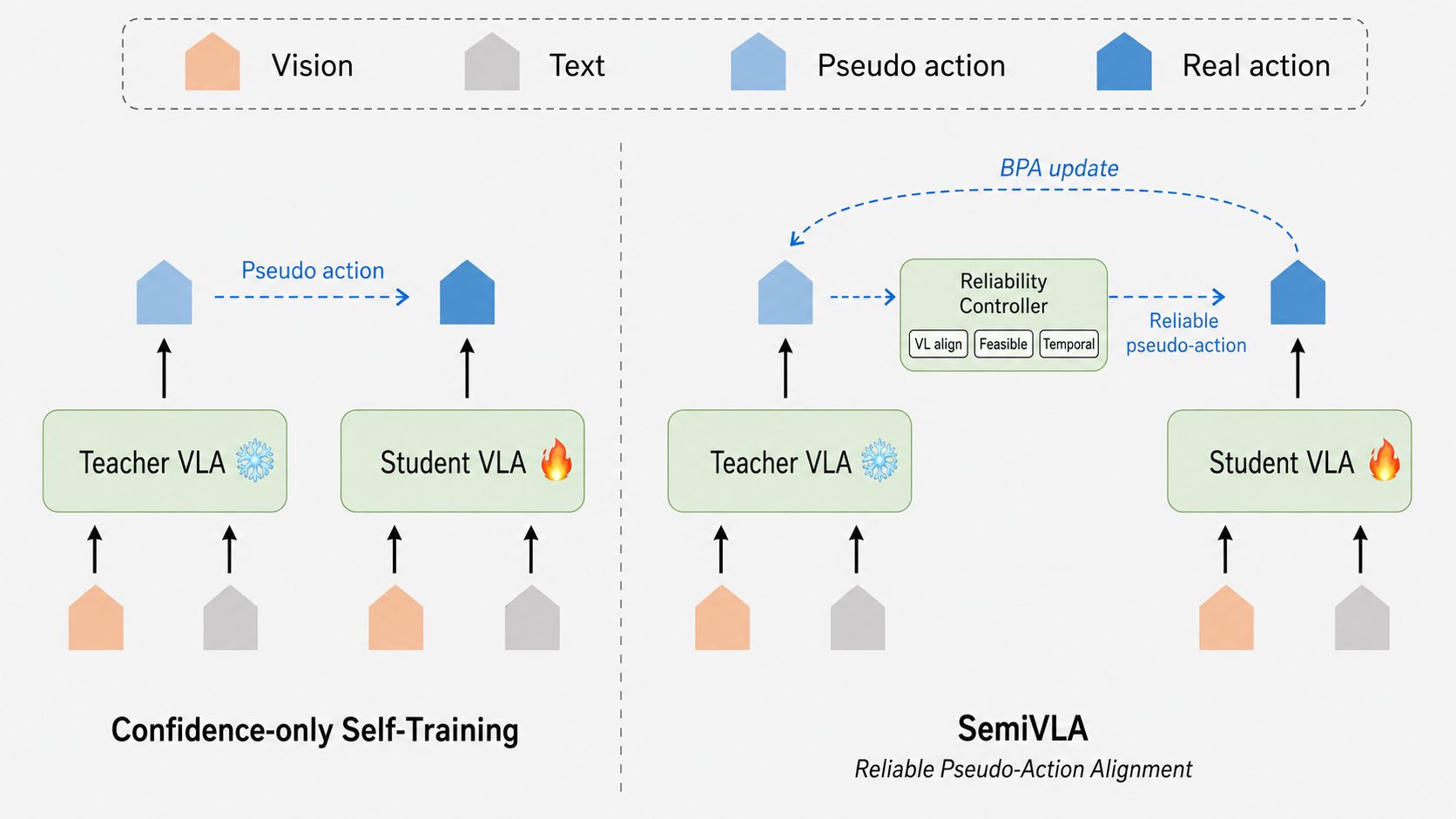}
    \caption{
    \textbf{Motivation and key difference between confidence-only self-training and SemiVLA.}
    Conventional confidence-only self-training directly transfers teacher-generated pseudo-actions to the student, which is unreliable for VLA adaptation because high-confidence actions may still be visually ungrounded, physically infeasible, or temporally inconsistent.
    SemiVLA instead selects reliable pseudo-actions through a VLA-specific reliability controller and updates the teacher through Bottleneck-Projected Alignment, reducing noisy feedback contamination under limited action supervision.
    }
    \label{fig:diff}
\end{figure}
Vision-Language-Action (VLA) models have become an important direction for general-purpose robotic manipulation. 
By integrating visual perception, language understanding, and action generation into a unified policy, VLA models can directly map visual observations and natural language instructions to executable robot actions. 
Recent models have shown strong potential in language-conditioned manipulation, long-horizon control, and cross-task generalization~\cite{brohan2023rt-1,zitkovich2023rt-2,team2024octo,kim2024openvla,black2024pi_0,bjorck2025gr00t}. 
This progress suggests that large-scale vision-language priors can provide useful semantic knowledge for embodied decision making. 
However, adapting VLA models to new robots, new environments, and new task distributions still heavily depends on action-labeled robot demonstrations~\cite{kim2024openvla,kim2025fine,mitra2025mechanistic}. 
Such demonstrations are expensive to collect because they require teleoperation, embodiment-specific action recording, and careful task execution. 
This creates a practical bottleneck for deploying VLA models in realistic robotic settings.

A natural way to reduce this cost is to exploit unlabeled or weakly labeled robot data. 
In many robotic scenarios, visual observations and language instructions are easier to obtain than dense action labels. 
For example, one may have access to robot videos, egocentric manipulation videos, or trajectories with task descriptions, but without reliable low-level action annotations~\cite{oneill2024open,DBLP:conf/rss/KhazatskyP0BDKN24,team2024octo}. 
This setting is closely related to semi-supervised learning (SSL), where a model learns from a small labeled set and a large unlabeled set. 
SSL has been widely studied in visual recognition, where pseudo-labeling, teacher-student learning, and consistency regularization are commonly used to improve learning under limited labels~\cite{tarvainen2017mean,xie2020unsupervised,sohn2020fixmatch,wang2022freematch}. 
However, directly applying conventional SSL to VLA adaptation is non-trivial. 
In image classification, a pseudo-label is usually a discrete class. 
In VLA adaptation, a pseudo-label is an action prediction, which can be a continuous control vector or a sequence of action tokens. 
This pseudo-action must be semantically consistent with the language instruction, visually grounded in the observation, physically feasible in the robot action space, and temporally stable along the trajectory.

In this paper, we study \textbf{semi-supervised VLA adaptation under limited supervision signals}. 
Given a small action-labeled trajectory set $\mathcal{D}_l=\{(\mathbf{o}_{1:T},q,\mathbf{a}_{1:T})\}$ and a large unlabeled vision-language trajectory set $\mathcal{D}_u=\{(\mathbf{o}_{1:T},q)\}$, the goal is to adapt a pretrained VLA model using both labeled and unlabeled data. 
The key supervision signal missing from $\mathcal{D}_u$ is the low-level action sequence. 
This protocol explicitly separates action-labeled and action-unlabeled supervision signals, as reflected in Table~\ref{tab:ssl_many_model_clean}. 
To the best of our knowledge, we are the first to explicitly formulate this problem from the limited-supervision-signal perspective for VLA model adaptation. 
This setting is different from standard supervised VLA fine-tuning, which assumes action labels are available for all training trajectories~\cite{kim2024openvla,kim2025fine,mitra2025mechanistic}. 
It is also different from conventional semi-supervised visual learning, because the missing supervision is not a class label but an embodied action signal tied to language, vision, and dynamics.

This problem raises three main challenges. 
First, pseudo-actions are more fragile than pseudo-labels. 
An incorrect pseudo-action may cause compounding errors in closed-loop execution, because a small action error can shift the robot into an unseen state. 
Second, model confidence is not sufficient to measure pseudo-action quality. 
A VLA model may assign high confidence to an action that is visually ungrounded or physically infeasible. 
Third, teacher-student self-training can suffer from feedback contamination. 
If noisy student updates are directly averaged into the teacher, the teacher may become a worse pseudo-action generator and further amplify errors during self-distillation~\cite{tarvainen2017mean}. 
Therefore, semi-supervised VLA adaptation requires reliability estimation and feedback control mechanisms that are specific to vision-language-action learning.

To address these challenges, we propose \textbf{SemiVLA}, a self-distilled semi-supervised adaptation framework for VLA models. 
SemiVLA first performs supervised warm-up on the limited labeled trajectories, so that the student policy obtains a stable adaptation direction for the target action space. 
It then initializes a teacher policy from the warmed-up student and uses the teacher to generate pseudo-actions on unlabeled vision-language trajectories. 
The student learns from these pseudo-actions under stronger visual and temporal perturbations. 
To avoid learning from unreliable pseudo-actions, SemiVLA introduces a VLA-specific reliability controller that evaluates pseudo-actions from three complementary aspects: vision-language alignment, action feasibility, and temporal transition consistency. 
Only pseudo-actions that are semantically grounded, physically feasible, and temporally consistent are used for self-distillation.

\begin{figure*}[t]
    \centering
    \includegraphics[width=\linewidth]{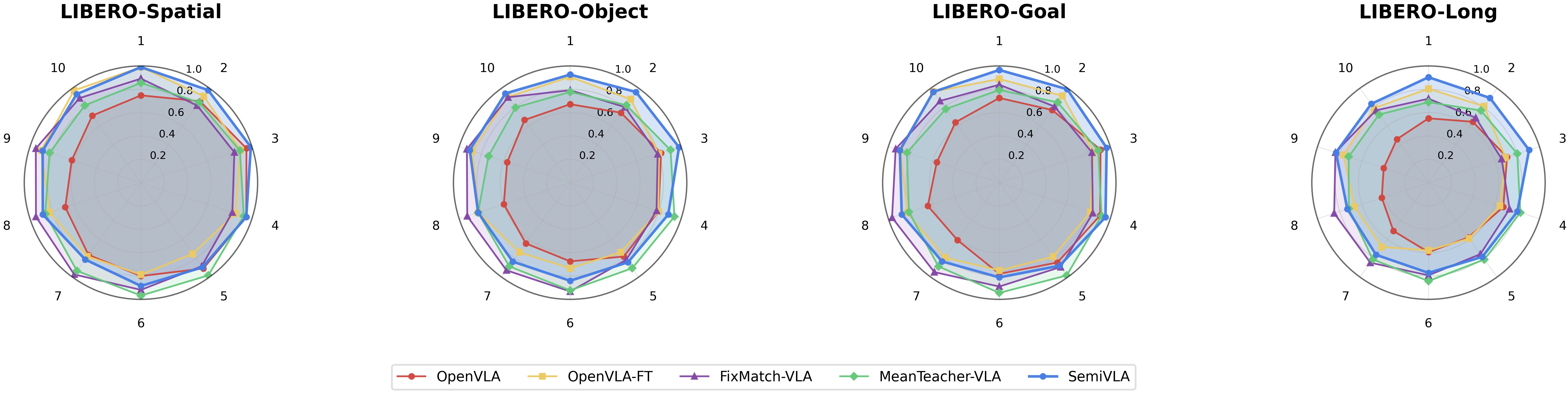}
    \caption{
    \textbf{Task-wise success rates across LIBERO suites.}
    We visualize per-task success rates on LIBERO-Spatial, LIBERO-Object, LIBERO-Goal, and LIBERO-Long.
    Compared with zero-shot OpenVLA, supervised fine-tuning, and conventional semi-supervised baselines, SemiVLA achieves more consistent task-wise improvements across all four suites.
    The gains are especially visible on LIBERO-Long, where reliable pseudo-action selection and temporal transition consistency help reduce failure accumulation in long-horizon manipulation.
    }
    \label{fig:radar_libero}
\end{figure*}

SemiVLA further introduces a Bottleneck-Projected Alignment Update for teacher evolution, motivated by the information bottleneck principle~\cite{tishby2000information,DBLP:conf/iclr/AlemiFD017}. 
Instead of updating the teacher by a conventional scalar EMA~\cite{tarvainen2017mean}, SemiVLA filters the student-to-teacher update direction before transferring it to the teacher. 
The update is decomposed into a semantic alignment path and an action precision path. 
The semantic path transfers only bottleneck-stable vision-language updates, while the action path transfers only updates supported by feasible and temporally consistent pseudo-actions. 
This design prevents nuisance-sensitive visual updates and unstable action corrections from contaminating the teacher. 
As a result, the teacher becomes a more reliable pseudo-action generator, and the student-teacher loop evolves through vision-language-action aligned feedback. 
Figure~\ref{fig:main_results_visual} visualizes the benchmark-level trend. 
SemiVLA improves both the average success rate and the task-wise results, especially on long-horizon and cross-benchmark settings. 
The task-wise radar visualization is provided in Fig.~\ref{fig:radar_libero}, showing that SemiVLA improves consistency across individual LIBERO tasks.

Our contributions are summarized as follows:
\begin{itemize}
    \item We introduce a new problem setting, \textbf{semi-supervised VLA adaptation under limited supervision signals}, where only a small subset of trajectories contains action labels and the remaining trajectories provide vision-language observations without action supervision.
    \item We propose \textbf{SemiVLA}, a self-distilled teacher-student framework that adapts pretrained VLA models from limited action supervision and unlabeled vision-language trajectories.
    \item We design a VLA-specific reliability controller that evaluates pseudo-actions through vision-language alignment, action feasibility, and temporal transition consistency, making pseudo-action learning more suitable for embodied control than confidence-only SSL.
    \item We propose a Bottleneck-Projected Alignment Update that filters student feedback before updating the teacher, reducing teacher contamination and improving the reliability of self-distilled pseudo-actions.
    \item We conduct experiments on language-conditioned robotic manipulation benchmarks under limited action-label settings, and show that SemiVLA improves adaptation performance over supervised fine-tuning and conventional pseudo-action self-training.
\end{itemize}

\section{Related Work and Preliminaries}
\label{sec:preliminaries}

\subsection{Vision-Language-Action Models}

Vision-Language-Action (VLA) models extend vision-language foundation models to embodied control by directly predicting robot actions from visual observations and language instructions. 
Early large-scale robot policies such as RT-1 and RT-2 demonstrate that transformer-based models can learn language-conditioned manipulation from diverse robot demonstrations~\cite{brohan2023rt-1,zitkovich2023rt-2}. 
More recent VLA models, including OpenVLA, $\pi_0$, Octo, GR00T, and related generalist robot policies, further improve cross-task generalization, language grounding, and action generation by leveraging large-scale robot datasets and pretrained vision-language backbones~\cite{kim2024openvla,black2024pi_0,team2024octo,bjorck2025gr00t}. 
These models typically formulate robot control as conditional action prediction, where the model receives an observation $\mathbf{o}_t$ and a language instruction $q$, and outputs an action $\mathbf{a}_t$ or an action chunk $\mathbf{a}_{t:t+H}$.

Despite their strong generalization ability, VLA models still require task-specific adaptation when deployed on new robots, new environments, or new task distributions. 
This adaptation is usually performed with action-labeled demonstrations~\cite{kim2024openvla,kim2025fine,mitra2025mechanistic}. 
However, action-labeled demonstrations are expensive because they require physical robot execution, teleoperation, embodiment-specific action recording, and careful task verification. 
This creates a gap between the availability of unlabeled visual-language trajectories and the scarcity of low-level robot action labels. 
SemiVLA targets this gap by studying semi-supervised VLA adaptation, where action labels are available only for a small subset of trajectories.

\subsection{VLA Fine-Tuning and Parameter-Efficient Adaptation}

Fine-tuning is an important step for adapting pretrained VLA models to downstream robot tasks. 
OpenVLA demonstrates that generalist VLA policies can be adapted to new robot setups with parameter-efficient methods such as LoRA and quantization~\cite{kim2024openvla}. 
OpenVLA-OFT further studies key fine-tuning design choices, including parallel decoding, action chunking, continuous action representation, and regression-based objectives, showing that better adaptation recipes can substantially improve both inference speed and task success~\cite{kim2025fine}. 
Other works explore lightweight adaptation modules, task-specific policy heads, selective parameter tuning, or mechanistic fine-tuning to reduce the cost of adapting large VLA backbones~\cite{wang2025vla-adapter,mitra2025mechanistic}.

These works mainly focus on how to better fine-tune VLA models when action-labeled demonstrations are available. 
In contrast, SemiVLA focuses on a complementary question: how can a VLA model exploit action-unlabeled vision-language trajectories when only a small portion of trajectories contains low-level action labels? 
This difference is important because simply reducing the number of trainable parameters does not solve the missing-action-supervision problem. 
A parameter-efficient method can make adaptation cheaper, but it still needs reliable action labels for each training trajectory. 
SemiVLA is compatible with PEFT methods such as Adapter, LoRA, QLoRA, and Selective LoRA, but its core contribution is to convert action-unlabeled trajectories into reliable pseudo-action supervision.

\subsection{Semi-Supervised Learning}

Semi-supervised learning (SSL) aims to improve model training by using a small labeled set and a large unlabeled set. 
A common paradigm is pseudo-labeling, where a teacher or the current model generates labels for unlabeled samples, and the student is trained to match these pseudo-labels. 
Consistency regularization is another widely used strategy, where predictions are encouraged to remain stable under perturbations of the same input~\cite{xie2020unsupervised,sohn2020fixmatch,wang2022freematch}. 
Teacher-student frameworks such as Mean Teacher update the teacher from the student and use the teacher as a stable pseudo-label generator~\cite{tarvainen2017mean}. 
These approaches have been highly effective for visual classification and recognition tasks.

However, directly applying conventional SSL to VLA adaptation is not straightforward. 
In standard image classification, the pseudo-label is usually a discrete semantic class. 
The main question is whether the predicted class is correct. 
In VLA adaptation, the pseudo-label is an embodied action signal, which may be a continuous action vector, an action token sequence, or a multi-step action chunk. 
A pseudo-action must satisfy multiple constraints simultaneously: it must follow the language instruction, be grounded in the visual observation, obey robot action feasibility, and remain temporally consistent with the observed trajectory. 
Therefore, confidence-only pseudo-labeling is insufficient for VLA adaptation. 
SemiVLA addresses this issue by introducing a VLA-specific reliability controller that evaluates pseudo-actions through vision-language alignment, action feasibility, and temporal transition consistency.

\subsection{Learning from Unlabeled Videos and Latent Actions}

A related line of work studies how to exploit unlabeled videos for robot learning. 
Large-scale robot datasets and in-the-wild manipulation trajectories provide diverse visual dynamics that can support robot policy learning beyond fully labeled demonstrations~\cite{oneill2024open,DBLP:conf/rss/KhazatskyP0BDKN24}. 
Latent Action Models (LAMs) learn compact latent action representations from visual dynamics, usually by predicting future frames, modeling state transitions, or tokenizing visual motion without access to robot actions~\cite{ye2025latent,chen2025moto}. 
These latent actions can provide additional supervision for VLA learning, especially when robot action labels are scarce. 
Recent methods further align latent action representations with VLA policy features or jointly optimize latent action models and action-generation policies, showing that unlabeled videos can help regularize action learning~\cite{liu2026lara}.

These works are closely related to SemiVLA because both aim to reduce dependence on dense action-labeled robot datasets. 
However, their supervision form and training objective are different. 
LAM-based approaches usually learn an intermediate latent action space from visual dynamics and use it as an auxiliary representation or pseudo-label. 
SemiVLA instead studies semi-supervised VLA adaptation directly: the unlabeled data are vision-language trajectories without low-level action labels, and the goal is to generate reliable pseudo-actions for adapting the VLA policy itself. 
Moreover, SemiVLA does not rely on a separate latent action tokenizer. 
It uses a teacher VLA policy to generate pseudo-actions and controls their quality with explicit reliability estimation and feedback-controlled teacher evolution.

\subsection{Preliminaries: Semi-Supervised VLA Adaptation}

We consider a pretrained VLA policy $\pi$ that maps a visual observation $\mathbf{o}_t$ and a language instruction $q$ to a robot action $\mathbf{a}_t$. 
The action can be represented as continuous control values or discrete action tokens, depending on the underlying VLA architecture. 
In supervised VLA fine-tuning, every training trajectory contains action labels. 
Given a labeled trajectory set $\mathcal{D}_l=\{(\mathbf{o}_{i,1:T_i}, q_i, \mathbf{a}_{i,1:T_i})\}_{i=1}^{N_l}$, the standard fine-tuning objective minimizes an action prediction loss:
\begin{equation}
\mathcal{L}_{\mathrm{sup}}
=
\mathbb{E}_{(\mathbf{o}_t,q,\mathbf{a}_t)\sim\mathcal{D}_l}
\left[
\ell_{\mathrm{act}}
\left(
\pi(\mathbf{o}_t,q),
\mathbf{a}_t
\right)
\right],
\end{equation}
where $\ell_{\mathrm{act}}$ can be cross entropy for action-token prediction or regression loss for continuous action prediction.

In this paper, we study a semi-supervised adaptation setting. 
Besides the labeled set $\mathcal{D}_l$, we are given an unlabeled vision-language trajectory set $\mathcal{D}_u=\{(\mathbf{o}_{j,1:T_j},q_j)\}_{j=1}^{N_u}$, where low-level action labels are unavailable. 
The goal is to adapt the VLA policy using both $\mathcal{D}_l$ and $\mathcal{D}_u$. 
This setting differs from standard supervised fine-tuning because most trajectories do not provide action labels. 
It also differs from conventional SSL because the missing supervision is not a semantic class but an embodied action sequence.

A direct solution is to use a teacher policy to generate pseudo-actions for $\mathcal{D}_u$ and train a student policy on these pseudo-actions. 
However, pseudo-actions can be noisy even when the teacher is confident. 
They may be visually ungrounded, violate action feasibility, or fail to explain the temporal transition between consecutive observations. 
Therefore, SemiVLA formulates semi-supervised VLA adaptation as reliable pseudo-action learning. 
It introduces a teacher-student self-distillation framework, a VLA-specific reliability controller, and a Bottleneck-Projected Alignment Update to make pseudo-action supervision more reliable under limited action labels.
\section{Method}
\label{sec:method}

\subsection{Overview of SemiVLA}

\begin{figure*}[t]
    \centering
    \includegraphics[width=\linewidth]{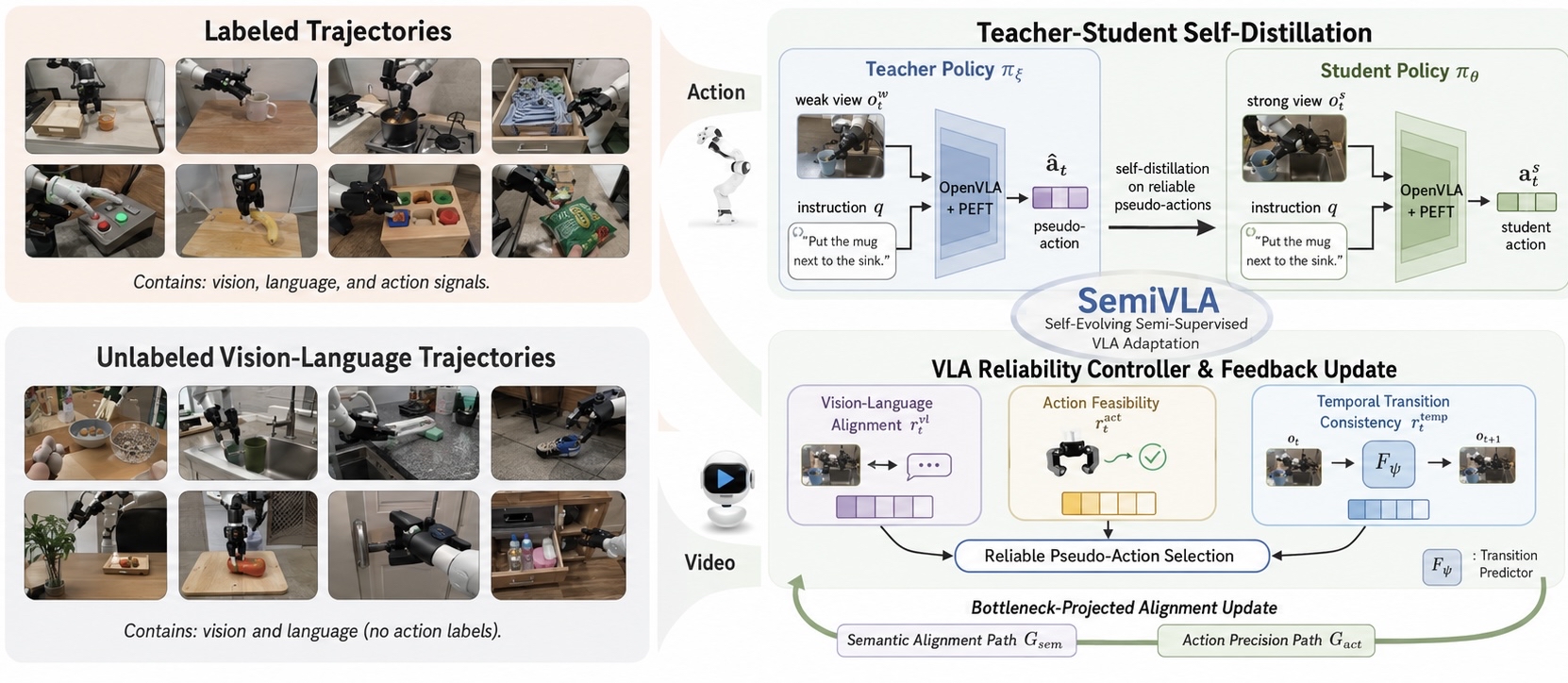}
    \caption{\textbf{Overview of SemiVLA.}}
    \label{fig:overview}
\end{figure*}

We propose \textbf{SemiVLA}, a self-distilled semi-supervised adaptation framework for Vision-Language-Action (VLA) models under limited supervision signals. 
SemiVLA contains two interacting VLA policies: a trainable student policy and a feedback-updated teacher policy. 
The student is first adapted with a small number of action-labeled trajectories, so that it obtains a stable initial understanding of the target robot embodiment, action space, and task distribution. 
The teacher is then initialized from the warmed-up student and generates pseudo-actions for unlabeled vision-language trajectories. 
The student learns from these pseudo-actions under stronger visual and temporal perturbations, while the teacher is updated only through bottleneck-projected alignment feedback from the student. 
This creates an asymmetric optimization path: the student evolves by gradient back-propagation, while the teacher evolves by alignment-filtered self-distillation feedback.

The main challenge is that VLA pseudo-labels are not ordinary class labels. 
They are action predictions that must be consistent with the language instruction, grounded in the visual observation, feasible in the robot action space, and stable across time. 
Therefore, SemiVLA introduces a VLA-specific reliability controller to estimate whether a pseudo-action is trustworthy. 
The controller evaluates each pseudo-action from three aspects: vision-language alignment, action feasibility, and temporal transition consistency. 
Only reliable pseudo-actions are used for self-distillation. 
Moreover, SemiVLA does not directly average all student parameters into the teacher. 
Instead, it transfers only the student update directions that are supported by stable visual-language subspaces and action-consistent temporal evidence. 
In this way, SemiVLA enables VLA models to adapt from limited action supervision and unlabeled trajectories through a self-evolving teacher-student process.

\subsection{Problem Setup and Supervised VLA Warm-up}

Let $\mathcal{D}_l=\{(\mathbf{o}_{i,1:T_i}, q_i, \mathbf{a}_{i,1:T_i})\}_{i=1}^{N_l}$ be the labeled trajectory set, where $\mathbf{o}_{i,t}$ denotes the visual observation at time step $t$, $q_i$ denotes the language instruction, and $\mathbf{a}_{i,t}$ denotes the ground-truth robot action. 
Let $\mathcal{D}_u=\{(\mathbf{o}_{j,1:T_j}, q_j)\}_{j=1}^{N_u}$ be the unlabeled trajectory set, where the action supervision is unavailable. 
SemiVLA adapts a pretrained VLA policy with limited action labels by training a student policy $\pi_\theta$ and a feedback teacher policy $\pi_\xi$. 
Both policies share the same pretrained initialization, and only lightweight adaptation parameters, such as LoRA or other PEFT modules, are updated.

SemiVLA first performs supervised warm-up on the limited labeled trajectories. 
For a labeled sample $(\mathbf{o}_t,q,\mathbf{a}_t)$, the student predicts $\pi_\theta(\mathbf{o}_t,q)$. 
We use a general action loss $\ell_{\mathrm{act}}$, which can be action-token cross entropy for discrete VLA action tokens or regression loss for continuous actions. 
The supervised warm-up objective is
\begin{equation}
\mathcal{L}_{\mathrm{sup}}
=
\frac{1}{|\mathcal{B}_l|}
\sum_{(\mathbf{o}_t,q,\mathbf{a}_t)\in\mathcal{B}_l}
\ell_{\mathrm{act}}
\left(
\pi_\theta(\mathbf{o}_t,q),
\mathbf{a}_t
\right).
\end{equation}
This stage gives the student a stable adaptation direction before unlabeled pseudo-action learning. 
After the warm-up stage, the teacher is initialized from the student by $\xi\leftarrow\theta$. 
This avoids generating pseudo-actions from a policy that has not yet learned the target action format.

\subsection{Self-Distilled Pseudo-Action Learning}

For an unlabeled trajectory $(\mathbf{o}_{1:T},q)$, SemiVLA constructs a weak view $\mathbf{o}_t^w$ and a strong view $\mathbf{o}_t^s$ for each time step. 
The weak view preserves the original task-relevant visual content, while the strong view applies stronger perturbations such as random crop, color jitter, camera noise, temporal jitter, or mild viewpoint disturbance. 
The teacher predicts on the weak view and produces a detached pseudo-action $\hat{\mathbf{a}}_t=\mathrm{sg}(\pi_\xi(\mathbf{o}_t^w,q))$. 
The student predicts on the strong view and produces $\mathbf{a}_t^s=\pi_\theta(\mathbf{o}_t^s,q)$. 
The student is then trained to match the teacher pseudo-action only when the pseudo-action is considered reliable.

Let $r_t$ denote the reliability score of the teacher pseudo-action at time step $t$, and let $M_t=\mathbb{I}[r_t\geq\tau_r]$ be the valid pseudo-action mask. 
The self-distillation loss is
\begin{equation}
\mathcal{L}_{\mathrm{sd}}
=
\frac{1}{|\mathcal{B}_u|}
\sum_{(\mathbf{o}_{1:T},q)\in\mathcal{B}_u}
\frac{1}{T}
\sum_{t=1}^{T}
M_t r_t
\cdot
d
\left(
\pi_\theta(\mathbf{o}_t^s,q),
\mathrm{sg}(\pi_\xi(\mathbf{o}_t^w,q))
\right),
\end{equation}
where $d(\cdot,\cdot)$ is KL divergence or cross entropy for action-token prediction, and $L_1$ or $L_2$ distance for continuous action prediction. 
This objective can be viewed as a VLA version of weak-strong self-training. 
However, SemiVLA does not trust teacher confidence alone. 
A high-confidence action can still be visually ungrounded, physically infeasible, or temporally inconsistent. 
Therefore, the reliability score $r_t$ is computed by a VLA-specific reliability controller.

\subsection{Vision-Language-Action Reliability Controller}

The reliability controller evaluates each pseudo-action from three complementary signals: vision-language alignment, action feasibility, and temporal transition consistency. 
For an unlabeled step $(\mathbf{o}_t,q)$, SemiVLA defines the pseudo-action reliability as
\begin{equation}
r_t
=
r_t^{\mathrm{vl}}
\cdot
r_t^{\mathrm{act}}
\cdot
r_t^{\mathrm{temp}}.
\end{equation}
Here, $r_t^{\mathrm{vl}}$ measures whether the visual observation contains task-relevant evidence for the language instruction, $r_t^{\mathrm{act}}$ measures whether the predicted action is feasible and stable, and $r_t^{\mathrm{temp}}$ measures whether the pseudo-action is consistent with the observed temporal transition. 
The multiplicative form is deliberately strict. 
If any part is unreliable, the final pseudo-action weight is reduced.

\textbf{Vision-language alignment.}
The vision-language score checks whether the instruction is grounded in the current observation. 
Let $z_q$ be the language embedding of the instruction and $z_o$ be the visual embedding of the observation. 
The score is computed as $r_t^{\mathrm{vl}}=\sigma(\mathrm{sim}(z_q,z_o)/\tau_{\mathrm{vl}})$, where $\mathrm{sim}(\cdot,\cdot)$ denotes cosine similarity and $\tau_{\mathrm{vl}}$ is a temperature. 
When attention maps or visual token scores are available, $z_o$ is obtained from instruction-relevant visual tokens rather than the global image feature. 
This encourages the pseudo-action to be supported by the correct object or region in the scene. 
For example, when the instruction is to place a red mug on a plate, the pseudo-action should be grounded in the visual tokens corresponding to the red mug and the plate.

\textbf{Action feasibility.}
The action feasibility score rejects pseudo-actions that are unstable or physically implausible. 
For continuous actions, we measure action magnitude, temporal variation, and gripper stability. 
For action-token VLAs, we measure the entropy of the action-token distribution and the consistency of decoded actions across neighboring frames. 
A simple continuous-action feasibility score is defined as $r_t^{\mathrm{act}}=\exp(-\|\hat{\mathbf{a}}_t-\hat{\mathbf{a}}_{t-1}\|_1/\tau_a)\cdot\mathbb{I}[\|\hat{\mathbf{a}}_t\|_\infty<\delta_a]$, where $\tau_a$ controls smoothness sensitivity and $\delta_a$ bounds abnormal action magnitude. 
This term prevents the student from learning pseudo-actions with sudden jumps, unrealistic end-effector movement, or unstable gripper behavior.

\textbf{Temporal transition consistency.}
A reliable pseudo-action should also explain the transition from the current observation to the next observation. 
SemiVLA uses a lightweight transition predictor $F_\psi$ in the representation space. 
Let $h(\cdot)$ denote a frozen visual encoder. 
Given $h(\mathbf{o}_t)$, the language embedding $z_q$, and the teacher pseudo-action $\hat{\mathbf{a}}_t$, $F_\psi$ predicts the next visual state feature. 
The temporal transition loss is
\begin{equation}
\mathcal{L}_{\mathrm{trans}}
=
\frac{1}{|\mathcal{B}_u|}
\sum_{(\mathbf{o}_{1:T},q)\in\mathcal{B}_u}
\frac{1}{T-1}
\sum_{t=1}^{T-1}
M_t r_t
\left\|
h(\mathbf{o}_{t+1})
-
F_\psi
\left(
h(\mathbf{o}_t),
\hat{\mathbf{a}}_t,
z_q
\right)
\right\|_2^2.
\end{equation}
The temporal reliability $r_t^{\mathrm{temp}}$ is computed from the negative transition error. 
A pseudo-action receives a high temporal score when it can explain the observed visual transition. 
This design makes self-distillation action-aware rather than confidence-only. 
It also reduces the risk that the student learns actions that are semantically plausible but dynamically wrong.

\subsection{Vision-Language-Action Alignment}

Besides using reliability as a sample weight, SemiVLA further introduces an explicit alignment objective to connect visual grounding, language semantics, and action prediction. 
The alignment objective is based on the intuition that a valid VLA action should satisfy three conditions. 
First, the language instruction should identify the task-relevant visual evidence. 
Second, the predicted action should be compatible with that visual evidence. 
Third, the action should induce a plausible temporal transition. 
Therefore, we define the VLA alignment loss as
\begin{equation}
\mathcal{L}_{\mathrm{align}}
=
\frac{1}{|\mathcal{B}_u|}
\sum_{(\mathbf{o}_{1:T},q)\in\mathcal{B}_u}
\frac{1}{T}
\sum_{t=1}^{T}
M_t
\left[
(1-r_t^{\mathrm{vl}})
+
\lambda_a(1-r_t^{\mathrm{act}})
+
\lambda_t(1-r_t^{\mathrm{temp}})
\right],
\end{equation}
where $\lambda_a$ and $\lambda_t$ balance action feasibility and temporal consistency. 
This objective encourages the adaptation process to improve pseudo-action quality instead of merely increasing teacher confidence. 
As a result, the student learns from pseudo-actions that are better aligned with language, vision, and robot dynamics.

\subsection{Final Student Objective}

The student is trained by supervised imitation learning, self-distilled pseudo-action learning, VLA alignment, and temporal transition consistency. 
For unlabeled data, SemiVLA applies pseudo-action supervision only when the reliability mask is valid. 
The final student objective is
\begin{equation}
\mathcal{L}
=
\mathcal{L}_{\mathrm{sup}}
+
\lambda_{\mathrm{sd}} R_u(e)\mathcal{L}_{\mathrm{sd}}
+
\lambda_{\mathrm{align}}\mathcal{L}_{\mathrm{align}}
+
\lambda_{\mathrm{trans}}\mathcal{L}_{\mathrm{trans}},
\end{equation}
where $e$ is the current epoch and $R_u(e)$ is an unsupervised ramp-up schedule. 
The ramp-up schedule prevents noisy pseudo-actions from dominating early adaptation. 
The optimizer updates only the student PEFT parameters, the student action head, and the lightweight transition predictor. 
The teacher parameters are detached from gradient computation. 
Thus, the student is the only model optimized by back-propagation, while the teacher is updated only by bottleneck-projected alignment feedback.

\subsection{Bottleneck-Projected Alignment Update}

After each student update, SemiVLA updates the teacher through a \textbf{Bottleneck-Projected Alignment Update} rather than a conventional EMA step. 
A standard EMA transfers the student state to the teacher with a scalar averaging strength. 
This is insufficient for VLA adaptation, because the student update may contain nuisance-sensitive visual changes, unstable action corrections, or language-irrelevant policy drift. 
Directly averaging these changes into the teacher can corrupt the pseudo-action generator and amplify errors in later self-distillation. 
SemiVLA therefore updates the teacher only along parameter directions that are supported by stable vision-language-action evidence.

The key idea is to filter the \emph{student-to-teacher update direction} instead of only filtering input features. 
Let $\Delta=\theta-\xi$ denote the parameter displacement from the teacher to the current student. 
SemiVLA decomposes the trainable parameters into two groups: the semantic alignment group $\mathcal{G}_{\mathrm{sem}}$ and the action precision group $\mathcal{G}_{\mathrm{act}}$. 
The semantic alignment group contains visual projector and vision-language PEFT parameters. 
The action precision group contains the action head and late policy PEFT parameters. 
The two groups are updated with different alignment signals, because robust semantic grounding and precise action execution require different forms of reliability.

For the semantic alignment group, SemiVLA builds a bottleneck gate from weak and strong visual features. 
Let $A_t^w$ and $A_t^s$ denote the channel-wise gates computed from the weak and strong views, respectively. 
Each gate is obtained from the covariance structure of visual tokens and indicates which feature channels remain stable under perturbation. 
The bottleneck stability score is defined as $r_t^{\mathrm{ib}}=\exp(-\|A_t^w-A_t^s\|_F/\tau_{\mathrm{ib}})$, where $\tau_{\mathrm{ib}}$ controls sensitivity to view-induced feature changes. 
A high $r_t^{\mathrm{ib}}$ means that the visual subspace used by the policy is stable and less likely to be dominated by nuisance perturbations.

SemiVLA then computes group-specific update reliability. 
The semantic update reliability is $\omega_{\mathrm{sem}}=\overline{r^{\mathrm{ib}}r^{\mathrm{vl}}}$, while the action update reliability is $\omega_{\mathrm{act}}=\overline{r^{\mathrm{act}}r^{\mathrm{temp}}}$, where the overline denotes batch averaging. 
The semantic group is therefore updated only when the visual-language subspace is stable. 
The action group is updated only when pseudo-actions are feasible and temporally consistent. 
This dual-path design avoids a single robustness score dominating the whole teacher update. 
It also prevents semantic denoising from suppressing the fine-grained spatial and motion cues required for manipulation.

For each parameter group $g\in\{\mathcal{G}_{\mathrm{sem}},\mathcal{G}_{\mathrm{act}}\}$, SemiVLA computes an adaptive update strength:
\begin{equation}
\rho_g
=
\rho_{\max}(e)
\cdot
\sigma
\left(
w_1\omega_g
-
w_2 e_g
-
w_3 v_g
\right),
\end{equation}
where $\omega_g$ is the group-specific alignment reliability, $e_g$ is the teacher-student prediction deviation, and $v_g$ is an update volatility term. 
For action-token VLAs, $e_g$ is computed by KL divergence between teacher and student action-token distributions. 
For continuous-action VLAs, it is computed by action distance. 
The volatility term is defined as $v_g=\|\Delta_g-\bar{\Delta}_g\|_2/(\|\bar{\Delta}_g\|_2+\epsilon)$, where $\bar{\Delta}_g$ is the running average of previous update directions. 
This term suppresses sudden parameter changes caused by noisy pseudo-actions.

The teacher is updated by a bottleneck-projected displacement:
\begin{equation}
\xi_g
\leftarrow
\xi_g
+
\rho_g
\cdot
\mathcal{P}_g(\Delta_g),
\qquad
g\in\{\mathcal{G}_{\mathrm{sem}},\mathcal{G}_{\mathrm{act}}\}.
\end{equation}
Here, $\mathcal{P}_g(\cdot)$ is an alignment projection operator. 
For the semantic alignment group, $\mathcal{P}_g$ projects the student displacement through the batch-averaged bottleneck gate, so that only stable vision-language channels are transferred to the teacher. 
For the action precision group, $\mathcal{P}_g$ keeps the full action update direction but scales it with action feasibility and temporal transition reliability. 
This preserves fine-grained control information while rejecting unstable pseudo-action drift.

In practice, the projection operator is implemented in a lightweight way. 
For LoRA parameters, the bottleneck gate is reduced to a channel-level vector and applied to the LoRA update direction. 
For action-head parameters, the projection is implemented as action-reliability scaling. 
All frozen pretrained backbone parameters are kept unchanged. 
Thus, the teacher receives only alignment-supported student feedback rather than blindly averaging all student changes.

Compared with conventional EMA, the proposed update has two advantages. 
First, it prevents noise-sensitive visual updates from contaminating the teacher, because semantic parameters are updated only through bottleneck-stable channels. 
Second, it preserves manipulation precision, because action parameters are controlled by feasibility and temporal consistency rather than visual robustness alone. 
As a result, the teacher becomes a more reliable pseudo-action generator during self-distillation, and the student-teacher loop evolves through vision-language-action aligned feedback.

\subsection{Training Schedule and Inference}

SemiVLA uses a three-stage training schedule. 
In the first stage, the model uses only supervised imitation learning on limited labeled trajectories. 
The objective is $\mathcal{L}=\mathcal{L}_{\mathrm{sup}}$, and the teacher is not used. 
This stage gives the student a stable policy initialization for the target action space. 
In the second stage, SemiVLA enables self-distilled pseudo-action learning and VLA-specific reliability estimation, but the teacher update is still disabled. 
The teacher is fixed in this stage, so the student can learn from stable pseudo-actions while the reliability controller becomes calibrated. 
In the third stage, SemiVLA enables the Bottleneck-Projected Alignment Update. 
The teacher then evolves with the student through alignment-filtered feedback, and the full objective is optimized.

At inference time, SemiVLA uses only the student policy. 
Given a visual observation $\mathbf{o}_t$ and a language instruction $q$, the student predicts the action by $\hat{\mathbf{a}}_t=\pi_\theta(\mathbf{o}_t,q)$. 
The teacher, reliability controller, transition predictor, and bottleneck-projected update module are not required during inference. 
Therefore, SemiVLA improves semi-supervised adaptation without adding extra inference cost.
\section{Experiments}
\label{sec:experiments}
\subsection{Results on Benchmark}

\begin{table*}[t]
\centering
\scriptsize
\renewcommand{\arraystretch}{1.12}
\setlength{\tabcolsep}{3.6pt}
\caption{
\textbf{Main results of SemiVLA on multiple VLA benchmarks under limited action supervision.}
All methods use OpenVLA as the backbone.
We report closed-loop success rate (\%).
``Sup.'' denotes the amount of action-labeled trajectories used for adaptation.
Higher is better.
}
\label{tab:main_results}
\resizebox{\textwidth}{!}{
\begin{tabular}{l c c c c c c c}
\toprule
\multirow{2}{*}{\textbf{Method}}
& \multirow{2}{*}{\textbf{Sup.}}
& \multirow{2}{*}{\textbf{Trainable Params}}
& \multicolumn{4}{c}{\textbf{LIBERO}}
& \multirow{2}{*}{\textbf{Avg.}} \\
\cmidrule(lr){4-7}
\rowcolor{headerblue}
& & & \textbf{Spatial} & \textbf{Object} & \textbf{Goal} & \textbf{Long} & \\
\midrule

\grouprow
\multicolumn{8}{l}{\textit{Zero-shot OpenVLA}} \\
\rowcolor{lightblue}
OpenVLA & 0 & 0 & 80.0 & 69.6 & 74.0 & 55.5 & 69.8 \\

\midrule
\grouprow
\multicolumn{8}{l}{\textit{Supervised adaptation with limited action labels}} \\
OpenVLA + Linear Probe & 10\% & 15.2M & 81.3 & 72.4 & 76.8 & 59.6 & 72.5 \\
OpenVLA + Adapter & 10\% & 42.7M & 84.6 & 78.2 & 81.5 & 65.7 & 77.5 \\
OpenVLA + LoRA & 10\% & 79.4M & 86.8 & 82.1 & 84.9 & 70.3 & 81.0 \\
OpenVLA + QLoRA & 10\% & 79.4M & 85.9 & 80.5 & 83.7 & 68.9 & 79.8 \\
OpenVLA + Full FT & 10\% & 7.0B & 84.5 & 79.8 & 82.6 & 66.8 & 78.4 \\

\midrule
\grouprow
\multicolumn{8}{l}{\textit{Semi-supervised adaptation with unlabeled trajectories}} \\
\semirow
SemiVLA + Adapter & 10\% & 42.7M & 88.9 & 84.7 & 87.2 & 74.6 & 83.9 \\
\semirow
SemiVLA + LoRA & 10\% & 79.4M & 91.7 & 88.6 & 90.9 & 80.5 & 87.9 \\
\semirow
SemiVLA + QLoRA & 10\% & 79.4M & 90.8 & 87.4 & 89.7 & 78.9 & 86.7 \\
\semirow
SemiVLA + Selective LoRA & 10\% & 31.6M & \best{92.4} & \best{89.5} & \best{91.8} & \best{82.3} & \best{89.0} \\
\semirow
SemiVLA + Full FT & 10\% & 7.0B & 90.5 & 87.8 & 89.2 & 78.5 & 86.5 \\

\bottomrule
\end{tabular}
}
\end{table*}

\begin{table*}[t]
\centering
\scriptsize
\renewcommand{\arraystretch}{1.12}
\setlength{\tabcolsep}{3.4pt}
\caption{
\textbf{Cross-benchmark evaluation of SemiVLA.}
All methods use OpenVLA as the backbone and are adapted with the same amount of labeled trajectories.
We report success rate (\%) for LIBERO and average completed tasks for CALVIN.
Since CALVIN uses a different scale, the last column reports the average over LIBERO task suites only.
Higher is better.
}
\label{tab:cross_benchmark}
\resizebox{\textwidth}{!}{
\begin{tabular}{l c c c c c c c}
\toprule
\rowcolor{headerblue}
\textbf{Method}
& \textbf{Sup.}
& \textbf{LIBERO-Spatial}
& \textbf{LIBERO-Object}
& \textbf{LIBERO-Goal}
& \textbf{LIBERO-Long}
& \textbf{CALVIN}
& \textbf{LIBERO Avg.} \\
\midrule

\rowcolor{lightblue}
OpenVLA Zero-shot & 0 & 80.0 & 69.6 & 74.0 & 55.5 & 1.32 & 69.8 \\

\midrule
\grouprow
\multicolumn{8}{l}{\textit{Supervised adaptation}} \\
OpenVLA + LoRA & 10\% & 86.8 & 82.1 & 84.9 & 70.3 & 1.78 & 81.0 \\
OpenVLA + Adapter & 10\% & 84.6 & 78.2 & 81.5 & 65.7 & 1.61 & 77.5 \\
OpenVLA + QLoRA & 10\% & 85.9 & 80.5 & 83.7 & 68.9 & 1.70 & 79.8 \\
OpenVLA + Selective LoRA & 10\% & 87.6 & 83.4 & 85.6 & 71.8 & 1.86 & 82.1 \\

\midrule
\grouprow
\multicolumn{8}{l}{\textit{Semi-supervised adaptation}} \\
\semirow
SemiVLA + LoRA & 10\% & 91.7 & 88.6 & 90.9 & 80.5 & 2.42 & 87.9 \\
\semirow
SemiVLA + Adapter & 10\% & 88.9 & 84.7 & 87.2 & 74.6 & 2.13 & 83.9 \\
\semirow
SemiVLA + QLoRA & 10\% & 90.8 & 87.4 & 89.7 & 78.9 & 2.31 & 86.7 \\
\semirow
SemiVLA + Selective LoRA & 10\% & \best{92.4} & \best{89.5} & \best{91.8} & \best{82.3} & \best{2.58} & \best{89.0} \\

\bottomrule
\end{tabular}
}
\end{table*}

\begin{table*}[t]
\centering
\scriptsize
\renewcommand{\arraystretch}{1.10}
\setlength{\tabcolsep}{2.4pt}
\caption{
\textbf{Many-model comparison under the proposed semi-supervised VLA adaptation setting.}
We evaluate all methods under a limited-supervision protocol, where only $10\%$ of target trajectories contain action labels, denoted as $\mathcal{D}_l$, and the remaining $90\%$ trajectories are action-unlabeled vision-language trajectories, denoted as $\mathcal{D}_u$.
Supervised baselines use only $\mathcal{D}_l$ and ignore $\mathcal{D}_u$, while SSL and latent-action baselines exploit $\mathcal{D}_u$ with their own mechanisms.
We report success rate (\%) on LIBERO and SIMPLER-ENV.
Higher is better.
}
\label{tab:many_model_ssl}
\label{tab:ssl_many_model_clean}
\resizebox{\textwidth}{!}{
\begin{tabular}{l c c c c c c c c c c c}
\toprule
\multirow{2}{*}{\textbf{Method}}
& \multirow{2}{*}{\makecell{\textbf{Labeled}\\$\boldsymbol{\mathcal{D}_l}$}}
& \multirow{2}{*}{\makecell{\textbf{Unlabeled}\\$\boldsymbol{\mathcal{D}_u}$}}
& \multicolumn{5}{c}{\textbf{LIBERO}}
& \multicolumn{4}{c}{\textbf{SIMPLER-ENV}} \\
\cmidrule(lr){4-8}
\cmidrule(lr){9-12}

& & 
& \textbf{Spatial}
& \textbf{Object}
& \textbf{Goal}
& \textbf{Long}
& \textbf{Avg.}
& \textbf{Pick}
& \textbf{Move}
& \textbf{Drawer}
& \textbf{Avg.} \\
\midrule

\grouprow
\multicolumn{12}{l}{\textit{Zero-shot baseline}} \\
\rowcolor{lightblue}
OpenVLA Zero-shot 
& 0 & 0 
& 80.0 & 69.6 & 74.0 & 55.5 & 69.8 
& 16.3 & 46.2 & 35.6 & 32.7 \\

\midrule
\grouprow
\multicolumn{12}{l}{\textit{Limited-label supervised adaptation baselines}} \\
Octo-FT 
& 10\% & 0 
& 75.8 & 77.1 & 72.4 & 53.5 & 69.7 
& 19.5 & 24.3 & 21.0 & 21.6 \\

OpenVLA-FT 
& 10\% & 0 
& 86.8 & 82.1 & 84.9 & 70.3 & 81.0 
& 36.8 & 48.2 & 39.1 & 41.4 \\

OpenVLA-OFT 
& 10\% & 0 
& 88.4 & 86.2 & 87.9 & 75.0 & 84.4 
& 52.5 & 57.3 & 40.8 & 50.2 \\

$\pi_0$-FAST-FT 
& 10\% & 0 
& 90.1 & 88.8 & 88.5 & 76.8 & 86.1 
& 58.4 & 63.1 & 45.9 & 55.8 \\

GR00T-N1.6-FT 
& 10\% & 0 
& 91.4 & 89.7 & 90.5 & 79.0 & 87.7 
& 70.2 & 74.5 & 54.6 & 66.4 \\

\midrule
\grouprow
\multicolumn{12}{l}{\textit{Semi-supervised and latent-action baselines}} \\
PseudoAction-VLA 
& 10\% & 90\% 
& 88.0 & 84.0 & 86.2 & 74.5 & 83.2 
& 44.0 & 55.2 & 38.1 & 45.8 \\

MeanTeacher-VLA 
& 10\% & 90\% 
& 89.1 & 85.2 & 87.0 & 77.1 & 84.6 
& 46.3 & 57.1 & 41.5 & 48.3 \\

FixMatch-VLA 
& 10\% & 90\% 
& 89.4 & 86.0 & 88.2 & 76.5 & 85.0 
& 47.5 & 58.0 & 41.6 & 49.0 \\

LAM-Pseudo 
& 10\% & 90\% 
& 89.8 & 87.3 & 88.8 & 77.0 & 85.7 
& 57.0 & 66.2 & 42.5 & 55.2 \\

LARA-style Align 
& 10\% & 90\% 
& 90.3 & 87.9 & 89.5 & 77.6 & 86.3 
& 60.5 & 70.8 & 43.0 & 58.1 \\

\midrule
\grouprow
\multicolumn{12}{l}{\textit{SemiVLA under the proposed SSL setting}} \\
\semirow
SemiVLA-OpenVLA 
& 10\% & 90\% 
& 92.4 & 89.5 & 91.8 & 82.3 & 89.0 
& 66.0 & 78.2 & 43.0 & 62.4 \\

\semirow
SemiVLA-OpenVLA-OFT 
& 10\% & 90\% 
& 93.0 & 91.5 & 92.6 & 87.5 & 91.2 
& 72.0 & 84.4 & 48.5 & 68.3 \\

\semirow
SemiVLA-$\pi_0$-FAST 
& 10\% & 90\% 
& \second{94.2} & \second{92.0} & \second{93.0} & \second{88.7} & \second{92.0} 
& \second{76.5} & \second{86.0} & 48.2 & \second{70.2} \\

\semirow
SemiVLA-GR00T-N1.6 
& 10\% & 90\% 
& \best{96.0} & \best{94.8} & \best{95.2} & \best{90.4} & \best{94.1} 
& \best{92.0} & \best{87.6} & \best{52.5} & \best{77.4} \\

\bottomrule
\end{tabular}
}
\end{table*}
\textbf{Many-model comparison under the proposed SSL setting.}
Table~\ref{tab:many_model_ssl} evaluates representative VLA models under our semi-supervised VLA adaptation protocol.
The compared models cover open-source generalist robot policies and recent VLA adaptation recipes, including Octo, OpenVLA, OpenVLA-OFT, $\pi_0$-style action-generation models, and GR00T-N1.6~\cite{team2024octo,kim2024openvla,kim2025fine,black2024pi_0,bjorck2025gr00t}.
In this protocol, only $10\%$ of target trajectories provide action labels, while the remaining $90\%$ trajectories contain only visual observations and language instructions.
Therefore, supervised baselines are trained only with $\mathcal{D}_l$ and ignore $\mathcal{D}_u$, whereas SSL and latent-action baselines are allowed to exploit the action-unlabeled trajectories.
This protocol directly evaluates whether a method can use unlabeled vision-language trajectories to compensate for missing action supervision.

Under this setting, supervised VLA fine-tuning improves over zero-shot OpenVLA, but remains limited by the small labeled set.
For example, OpenVLA-FT~\cite{kim2024openvla} obtains 81.0\% average success on LIBERO, while stronger supervised backbones such as $\pi_0$-FAST-FT~\cite{black2024pi_0} and GR00T-N1.6-FT~\cite{bjorck2025gr00t} reach 86.1\% and 87.7\%, respectively.
However, these methods do not use $\mathcal{D}_u$, so their performance mainly depends on the quality of the pretrained backbone and the limited labeled trajectories.

Semi-supervised baselines obtain better results by exploiting $\mathcal{D}_u$.
PseudoAction-VLA, MeanTeacher-VLA, and FixMatch-VLA are adapted from classical pseudo-labeling, teacher-student learning, and consistency-regularized SSL frameworks~\cite{xie2020unsupervised,tarvainen2017mean,sohn2020fixmatch,wang2022freematch}.
They improve over supervised OpenVLA-FT, indicating that unlabeled trajectories contain useful adaptation signals.
LAM-Pseudo and LARA-style Align further improve the results by using latent action modeling or representation alignment from unlabeled dynamics~\cite{ye2025latent,chen2025moto,liu2026lara}.
Nevertheless, these baselines still lack a VLA-specific reliability mechanism that jointly considers language grounding, action feasibility, and temporal consistency.

SemiVLA achieves the best performance across all evaluated backbones.
With OpenVLA as the backbone, SemiVLA improves the LIBERO average from 81.0\% to 89.0\%.
When combined with stronger action-generation backbones, SemiVLA further reaches 91.2\% with OpenVLA-OFT~\cite{kim2025fine}, 92.0\% with $\pi_0$-FAST~\cite{black2024pi_0}, and 94.1\% with GR00T-N1.6~\cite{bjorck2025gr00t}.
The same trend holds on SIMPLER-ENV, where SemiVLA-GR00T-N1.6 obtains 77.4\% average success.
These results show that SemiVLA is not simply a stronger supervised fine-tuning recipe.
Instead, it provides a general semi-supervised adaptation mechanism that effectively converts action-unlabeled vision-language trajectories into reliable pseudo-action supervision.
\begin{table}[t]
\centering
\scriptsize
\renewcommand{\arraystretch}{1.08}

\begin{minipage}{0.48\textwidth}
\centering
\setlength{\tabcolsep}{2.6pt}
\caption{
\textbf{Effect of labeled trajectory ratio.}
Average success rate (\%) across LIBERO task suites.
}
\label{tab:label_ratio}
\resizebox{\linewidth}{!}{
\begin{tabular}{l c c c c c}
\toprule
\rowcolor{headerblue}
\textbf{Method} & \textbf{1\%} & \textbf{5\%} & \textbf{10\%} & \textbf{20\%} & \textbf{Full} \\
\midrule
\grouprow
\multicolumn{6}{l}{\textit{Supervised adaptation}} \\
OpenVLA + LoRA & 72.6 & 76.8 & 81.0 & 84.7 & 88.1 \\
OpenVLA + Adapter & 70.8 & 74.5 & 77.5 & 81.3 & 85.2 \\
OpenVLA + QLoRA & 72.0 & 76.0 & 79.8 & 83.6 & 87.4 \\
\midrule
\grouprow
\multicolumn{6}{l}{\textit{Semi-supervised adaptation}} \\
\semirow
SemiVLA + LoRA & \second{79.8} & \second{84.9} & \second{87.9} & \second{90.2} & \second{91.8} \\
\semirow
SemiVLA + Adapter & 77.2 & 81.3 & 83.9 & 86.4 & 88.0 \\
\semirow
SemiVLA + QLoRA & 78.9 & 83.6 & 86.7 & 89.1 & 90.7 \\
\semirow
SemiVLA + Selective LoRA & \best{81.4} & \best{86.2} & \best{89.0} & \best{91.1} & \best{92.4} \\
\bottomrule
\end{tabular}
}
\end{minipage}
\hfill
\begin{minipage}{0.48\textwidth}
\centering
\setlength{\tabcolsep}{2.8pt}
\caption{
\textbf{Few-shot VLA adaptation.}
Average success rate (\%) across selected tasks.
}
\label{tab:few_shot}
\resizebox{\linewidth}{!}{
\begin{tabular}{l c c c c}
\toprule
\rowcolor{headerblue}
\textbf{Method} & \textbf{1-shot} & \textbf{5-shot} & \textbf{10-shot} & \textbf{20-shot} \\
\midrule
\rowcolor{lightblue}
OpenVLA Zero-shot & 42.8 & 42.8 & 42.8 & 42.8 \\
\midrule
\grouprow
\multicolumn{5}{l}{\textit{Supervised adaptation}} \\
OpenVLA + LoRA & 47.5 & 59.8 & 67.4 & 75.2 \\
OpenVLA + Adapter & 45.9 & 56.7 & 64.1 & 71.6 \\
OpenVLA + Selective LoRA & 49.8 & 63.5 & 71.2 & 79.1 \\
\midrule
\grouprow
\multicolumn{5}{l}{\textit{Semi-supervised adaptation}} \\
\semirow
SemiVLA + LoRA & \second{56.2} & \second{69.4} & \second{78.6} & \second{84.5} \\
\semirow
SemiVLA + Adapter & 53.1 & 65.2 & 74.0 & 80.8 \\
\semirow
SemiVLA + Selective LoRA & \best{58.7} & \best{72.8} & \best{81.9} & \best{87.3} \\
\bottomrule
\end{tabular}
}
\end{minipage}

\vspace{-2mm}
\end{table}

\begin{table*}[t]
\centering
\scriptsize
\renewcommand{\arraystretch}{1.10}
\setlength{\tabcolsep}{3.4pt}
\caption{
\textbf{Efficiency comparison under limited action supervision.}
All methods use OpenVLA as the backbone and are trained with 10\% labeled trajectories.
We report trainable parameters, training time, GPU memory, and average success rate.
}
\label{tab:efficiency}
\resizebox{\textwidth}{!}{
\begin{tabular}{l c c c c c}
\toprule
\rowcolor{headerblue}
\textbf{Method}
& \textbf{Trainable Params}
& \textbf{Trainable Ratio}
& \textbf{Training Time}
& \textbf{GPU Memory}
& \textbf{Avg. Success} \\
\midrule
\grouprow
\multicolumn{6}{l}{\textit{Supervised adaptation}} \\
OpenVLA + Full FT & 7.0B & 100\% & 9.8h & 78.0GB & 78.4 \\
OpenVLA + Adapter & 42.7M & 0.61\% & 3.1h & 25.4GB & 77.5 \\
OpenVLA + LoRA & 79.4M & 1.13\% & 3.6h & 27.8GB & 81.0 \\
OpenVLA + QLoRA & 79.4M & 1.13\% & 4.2h & 18.6GB & 79.8 \\
OpenVLA + Selective LoRA & 31.6M & 0.45\% & 3.0h & 24.2GB & 82.1 \\
\midrule
\grouprow
\multicolumn{6}{l}{\textit{Semi-supervised adaptation}} \\
\semirow
SemiVLA + Adapter & 42.7M & 0.61\% & 4.0h & 27.1GB & 83.9 \\
\semirow
SemiVLA + LoRA & 79.4M & 1.13\% & 4.7h & 29.6GB & \second{87.9} \\
\semirow
SemiVLA + QLoRA & 79.4M & 1.13\% & 5.1h & \best{20.4GB} & 86.7 \\
\semirow
SemiVLA + Selective LoRA & \best{31.6M} & \best{0.45\%} & \second{4.2h} & 24.8GB & \best{89.0} \\
\bottomrule
\end{tabular}
}
\end{table*}

\begin{table}[t]
\centering
\scriptsize
\renewcommand{\arraystretch}{1.10}
\setlength{\tabcolsep}{3.5pt}
\caption{
\textbf{Ablation study of SemiVLA components.}
All variants use OpenVLA with LoRA adaptation under 10\% labeled trajectories.
We report average success rate (\%) across LIBERO task suites.
}
\label{tab:ablation_components}
\resizebox{\columnwidth}{!}{
\begin{tabular}{l c c c c}
\toprule
\rowcolor{headerblue}
\textbf{Method}
& \textbf{Self-Distill}
& \textbf{VLA Reliability}
& \textbf{BPA Update}
& \textbf{Avg.} \\
\midrule
Supervised LoRA & \xmark & \xmark & \xmark & 81.0 \\
Naive pseudo-action & \cmark & \xmark & \xmark & 83.2 \\
+ VLA reliability controller & \cmark & \cmark & \xmark & 86.1 \\
+ Bottleneck-projected update & \cmark & \cmark & \cmark & \second{87.6} \\
\midrule
\semirow
SemiVLA Full & \cmark & \cmark & \cmark & \best{87.9} \\
\bottomrule
\end{tabular}
}
\end{table}

\begin{table*}[t]
\centering
\scriptsize
\renewcommand{\arraystretch}{1.10}
\setlength{\tabcolsep}{3.2pt}
\caption{
\textbf{Ablation of the VLA-specific reliability controller.}
All variants use SemiVLA + LoRA with OpenVLA backbone.
We report pseudo-action acceptance ratio and average success rate.
A lower acceptance ratio can indicate stricter pseudo-action filtering.
}
\label{tab:ablation_reliability}
\resizebox{\textwidth}{!}{
\begin{tabular}{l c c c c c c}
\toprule
\rowcolor{headerblue}
\textbf{Reliability Design}
& $\mathbf{r^{\mathrm{conf}}}$
& $\mathbf{r^{\mathrm{vl}}}$
& $\mathbf{r^{\mathrm{act}}}$
& $\mathbf{r^{\mathrm{temp}}}$
& \textbf{Accept. Ratio}
& \textbf{Avg.} \\
\midrule
Confidence only & \cmark & \xmark & \xmark & \xmark & 52.8 & 83.2 \\
Vision-language only & \xmark & \cmark & \xmark & \xmark & 46.5 & 84.0 \\
Action feasibility only & \xmark & \xmark & \cmark & \xmark & 48.1 & 84.5 \\
Temporal consistency only & \xmark & \xmark & \xmark & \cmark & 44.7 & 84.8 \\
VL + Action & \xmark & \cmark & \cmark & \xmark & 39.6 & 86.0 \\
\semirow
VL + Action + Temporal & \xmark & \cmark & \cmark & \cmark & \best{36.8} & \best{87.9} \\
\bottomrule
\end{tabular}
}
\end{table*}

\begin{table}[t]
\centering
\scriptsize
\renewcommand{\arraystretch}{1.10}
\setlength{\tabcolsep}{3.2pt}
\caption{
\textbf{Effect of teacher update strategy.}
All variants use SemiVLA + LoRA with OpenVLA backbone.
We report pseudo-action error, pseudo-action acceptance ratio, and average success rate.
}
\label{tab:ablation_update}
\resizebox{\columnwidth}{!}{
\begin{tabular}{l c c c}
\toprule
\rowcolor{headerblue}
\textbf{Teacher Update}
& \textbf{Pseudo-action Error} $\downarrow$
& \textbf{Accept. Ratio}
& \textbf{Avg. Success} $\uparrow$ \\
\midrule
Frozen teacher & 0.184 & 31.7 & 85.2 \\
Standard EMA & 0.171 & 40.5 & 85.9 \\
Reliability-aware EMA & \second{0.154} & \second{42.3} & \second{86.8} \\
\semirow
Bottleneck-projected update & \best{0.137} & 36.8 & \best{87.9} \\
\bottomrule
\end{tabular}
}
\end{table}

\begin{table*}[t]
\centering
\scriptsize
\renewcommand{\arraystretch}{1.10}
\setlength{\tabcolsep}{3.0pt}
\caption{
\textbf{Generalization and robustness under limited supervision.}
All methods use OpenVLA as the backbone.
Models are adapted on source tasks and evaluated on unseen tasks or perturbed environments.
We report success rate (\%).
}
\label{tab:generalization_robustness}
\resizebox{\textwidth}{!}{
\begin{tabular}{l c c c c c c}
\toprule
\rowcolor{headerblue}
\textbf{Method}
& \textbf{Source Tasks}
& \textbf{Unseen Task}
& \textbf{Lighting Var.}
& \textbf{Object Var.}
& \textbf{Distractor}
& \textbf{Avg.} \\
\midrule
\rowcolor{lightblue}
OpenVLA Zero-shot & -- & 48.2 & 55.6 & 49.4 & 45.0 & 49.6 \\
\midrule
\grouprow
\multicolumn{7}{l}{\textit{Supervised adaptation}} \\
OpenVLA + LoRA & Source & 61.3 & 69.5 & 63.2 & 58.7 & 63.2 \\
OpenVLA + Adapter & Source & 57.8 & 65.1 & 60.3 & 55.9 & 59.8 \\
OpenVLA + Selective LoRA & Source & 64.7 & 72.4 & 67.5 & 62.1 & 66.7 \\
\midrule
\grouprow
\multicolumn{7}{l}{\textit{Semi-supervised adaptation}} \\
\semirow
SemiVLA + LoRA & Source + Unlabeled & \second{72.8} & \second{80.1} & \second{75.4} & \second{70.9} & \second{74.8} \\
\semirow
SemiVLA + Adapter & Source + Unlabeled & 68.3 & 75.9 & 71.6 & 66.8 & 70.7 \\
\semirow
SemiVLA + Selective LoRA & Source + Unlabeled & \best{75.6} & \best{83.4} & \best{78.2} & \best{73.9} & \best{77.8} \\
\bottomrule
\end{tabular}
}
\end{table*}

\textbf{Main comparison.}
Table~\ref{tab:main_results} reports the main results on the four LIBERO task suites under the 10\% labeled-trajectory setting. 
All methods use OpenVLA as the backbone, so the comparison focuses on the effect of adaptation strategy rather than backbone capacity. 
The zero-shot OpenVLA baseline obtains an average success rate of 69.8\%, showing that the pretrained VLA already provides a strong general manipulation prior. 
However, its performance drops clearly on LIBERO-Long, where the success rate is only 55.5\%. 
This indicates that long-horizon manipulation remains difficult without task-specific adaptation.

With limited supervised adaptation, OpenVLA + LoRA achieves the best average result among the supervised baselines, reaching 81.0\%. 
This is 11.2 points higher than zero-shot OpenVLA, confirming that even a small amount of action supervision can adapt OpenVLA to the target benchmark. 
However, different supervised fine-tuning strategies show different behavior. 
Adapter tuning is more parameter-efficient but obtains a lower average success rate of 77.5\%. 
QLoRA reduces deployment cost but slightly underperforms standard LoRA. 
Full fine-tuning reaches only 78.4\% under 10\% labeled data, which suggests that updating all parameters with limited demonstrations may overfit the small labeled set and damage pretrained VLA priors.

SemiVLA consistently improves over the corresponding supervised adaptation baselines. 
For example, SemiVLA + LoRA improves the average success rate from 81.0\% to 87.9\%, giving a gain of 6.9 points. 
SemiVLA + Adapter improves OpenVLA + Adapter from 77.5\% to 83.9\%, and SemiVLA + QLoRA improves OpenVLA + QLoRA from 79.8\% to 86.7\%. 
These gains show that the improvement does not depend on a specific PEFT method. 
Instead, it comes from the proposed semi-supervised VLA adaptation framework, which uses unlabeled trajectories through reliable pseudo-action learning.

The strongest result is achieved by SemiVLA + Selective LoRA, which obtains 89.0\% average success rate. 
It improves over the best supervised baseline, OpenVLA + LoRA, by 8.0 points, and improves over zero-shot OpenVLA by 19.2 points. 
The gain is especially large on LIBERO-Long, where SemiVLA + Selective LoRA reaches 82.3\%, compared with 70.3\% for OpenVLA + LoRA. 
This 12.0-point improvement indicates that SemiVLA is particularly useful for long-horizon tasks, where pseudo-action reliability and temporal transition consistency are more important.

\textbf{Cross-benchmark evaluation.}
Table~\ref{tab:cross_benchmark} further evaluates SemiVLA across LIBERO and CALVIN. 
Since CALVIN reports the average number of completed tasks rather than success rate, the last column reports only the average over LIBERO task suites. 
This avoids mixing metrics with different scales. 
The overall trend is consistent with Table~\ref{tab:main_results}. 
SemiVLA improves all supervised counterparts on LIBERO, and the improvement also transfers to CALVIN.

On CALVIN, OpenVLA zero-shot obtains 1.32 completed tasks. 
Supervised adaptation improves this result, with OpenVLA + Selective LoRA reaching 1.86. 
However, SemiVLA + Selective LoRA further improves the result to 2.58. 
This is a gain of 0.72 over OpenVLA + Selective LoRA and 1.26 over zero-shot OpenVLA. 
The improvement on CALVIN suggests that SemiVLA does not merely fit the LIBERO task distribution. 
Instead, the self-distillation and VLA-specific alignment mechanism also improve transfer to a different long-horizon manipulation benchmark.

Among all semi-supervised variants, SemiVLA + Selective LoRA gives the best results on both LIBERO and CALVIN. 
This suggests that selective adaptation and semi-supervised pseudo-action learning are complementary. 
Selective LoRA focuses the trainable parameters on task-relevant model components, while SemiVLA improves the supervision signal by using reliable unlabeled trajectories. 
Together, they produce a stronger and more efficient adaptation strategy.

\section{Ablation Study}
\label{sec:ablation}

\begin{figure*}[t]
    \centering
    \includegraphics[width=\linewidth]{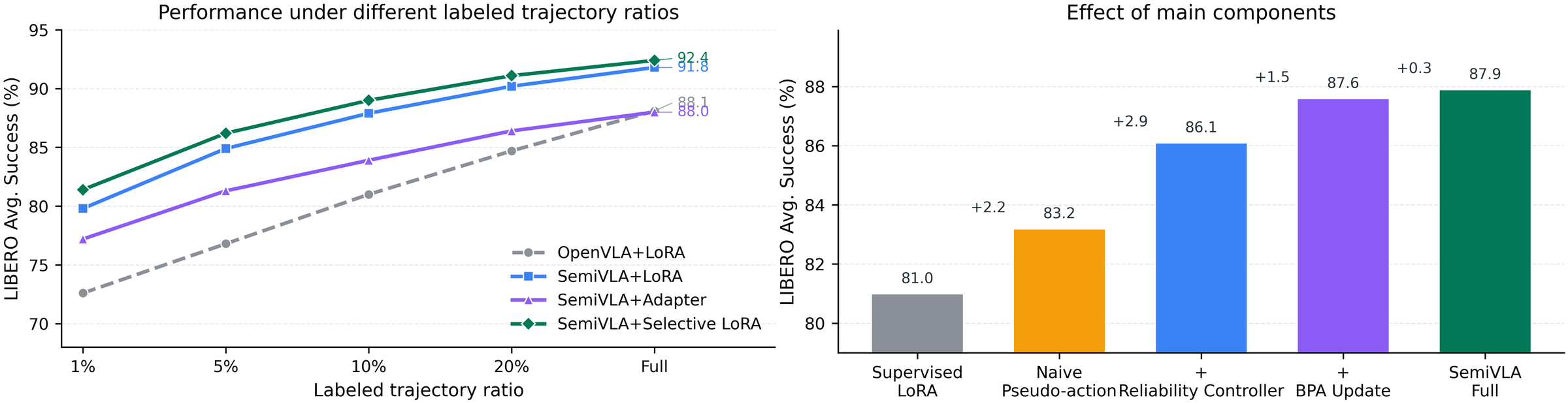}
    \caption{
    \textbf{Effect of labeled trajectory ratio and main SemiVLA components.}
    Left: performance under different labeled trajectory ratios.
    SemiVLA provides larger gains when action labels are scarce, confirming that unlabeled vision-language trajectories are most beneficial in the low-supervision regime.
    Right: component ablation of SemiVLA.
    Naive pseudo-action learning already improves over supervised LoRA, while the VLA reliability controller brings the largest gain by filtering unreliable pseudo-actions.
    The Bottleneck-Projected Alignment Update further improves performance by reducing noisy teacher feedback.
    }
    \label{fig:ablation_label_component}
\end{figure*}

As shown in Fig.~\ref{fig:ablation_label_component}, SemiVLA is particularly effective when the labeled trajectory ratio is low.
The component analysis further shows that the VLA reliability controller and Bottleneck-Projected Alignment Update are both necessary for reliable pseudo-action learning.

\textbf{Effect of labeled trajectory ratio.}
Table~\ref{tab:label_ratio} studies how different methods behave as the amount of labeled action supervision changes. 
The results show that SemiVLA is especially effective in the low-label regime. 
With only 1\% labeled trajectories, OpenVLA + LoRA achieves 72.6\%, while SemiVLA + LoRA reaches 79.8\%. 
SemiVLA + Selective LoRA further improves the result to 81.4\%. 
This shows that unlabeled trajectories provide a strong complementary supervision source when action labels are scarce.

The performance gap remains clear at 5\% and 10\% labeled ratios. 
At 10\%, OpenVLA + LoRA obtains 81.0\%, while SemiVLA + LoRA reaches 87.9\%. 
SemiVLA + Selective LoRA obtains 89.0\%, which is the best result in this setting. 
As the label ratio increases to 20\% and Full, all methods improve, but the relative gain from SemiVLA becomes smaller. 
This is expected because supervised action labels become more sufficient when the labeled set is large. 
However, SemiVLA + Selective LoRA still achieves the best result at Full supervision, reaching 92.4\%. 
This indicates that the proposed self-distillation framework also works as a regularizer, even when more labeled trajectories are available.

Overall, Table~\ref{tab:label_ratio} supports the central motivation of this paper. 
SemiVLA is designed for limited supervision signals, and the largest gains appear when labeled action trajectories are scarce. 
This confirms that semi-supervised VLA adaptation is a meaningful setting rather than a simple extension of standard supervised fine-tuning.

\textbf{Few-shot adaptation.}
Table~\ref{tab:few_shot} evaluates few-shot VLA adaptation with 1, 5, 10, and 20 demonstrations. 
OpenVLA zero-shot obtains 42.8\% average success rate, showing that the selected tasks require target-domain adaptation. 
Supervised fine-tuning improves performance as the number of demonstrations increases. 
For example, OpenVLA + LoRA improves from 47.5\% with 1-shot to 75.2\% with 20-shot. 
OpenVLA + Selective LoRA performs better than standard LoRA at all shot numbers, which suggests that task-specific parameter selection is useful when only a few demonstrations are available.

SemiVLA consistently outperforms the supervised variants across all few-shot settings. 
With 1-shot supervision, SemiVLA + Selective LoRA achieves 58.7\%, which is 8.9 points higher than OpenVLA + Selective LoRA and 11.2 points higher than OpenVLA + LoRA. 
With 20-shot supervision, SemiVLA + Selective LoRA reaches 87.3\%, outperforming OpenVLA + Selective LoRA by 8.2 points. 
The consistent gain shows that SemiVLA can use unlabeled trajectories to compensate for missing action labels.

The improvement is particularly important in the 1-shot and 5-shot settings. 
In these settings, supervised fine-tuning has very limited action supervision and is more likely to overfit. 
SemiVLA reduces this issue by using teacher-generated pseudo-actions and filtering them through the VLA-specific reliability controller. 
Thus, Table~\ref{tab:few_shot} shows that SemiVLA is not only effective under ratio-based label scarcity, but also under demonstration-based few-shot adaptation.

\textbf{Efficiency comparison.}
Table~\ref{tab:efficiency} compares different adaptation strategies in terms of trainable parameters, trainable ratio, training time, GPU memory, and average success rate. 
Full fine-tuning updates all 7.0B parameters and requires 78.0GB GPU memory, but under 10\% labeled trajectories it only obtains 78.4\% average success rate. 
This shows that large trainable capacity does not necessarily lead to better adaptation when supervision is limited. 
In contrast, parameter-efficient methods obtain better trade-offs between performance and cost.

Among supervised baselines, OpenVLA + LoRA reaches 81.0\% with 79.4M trainable parameters, while OpenVLA + Selective LoRA reaches 82.1\% with only 31.6M parameters. 
This indicates that selecting task-relevant parameters can improve both efficiency and performance. 
QLoRA has the lowest GPU memory among supervised methods, requiring 18.6GB, but its performance is slightly lower than standard LoRA. 
This suggests a trade-off between memory efficiency and adaptation quality.

SemiVLA improves the performance of all parameter-efficient strategies with moderate additional training cost. 
SemiVLA + LoRA improves average success from 81.0\% to 87.9\%, while increasing training time from 3.6h to 4.7h. 
SemiVLA + Selective LoRA achieves the best average success rate of 89.0\% with only 31.6M trainable parameters and 24.8GB GPU memory. 
This is a strong efficiency-performance trade-off. 
The result suggests that SemiVLA does not require full-parameter tuning to be effective. 
Instead, reliable pseudo-action learning and bottleneck-projected teacher feedback can improve adaptation within a small trainable parameter budget.

\begin{figure*}[t]
    \centering
    \includegraphics[width=\linewidth]{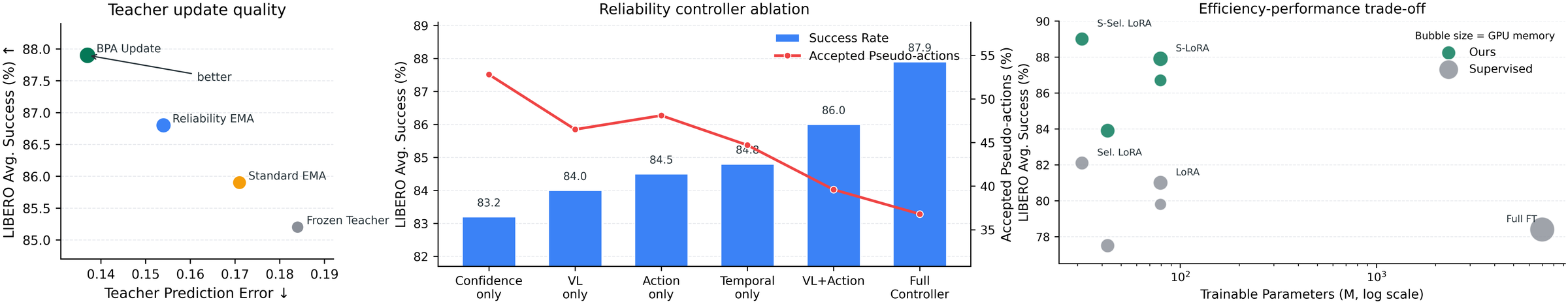}
    \caption{
    \textbf{Analysis of teacher evolution, reliability control, and efficiency-performance trade-off.}
    Left: teacher update quality measured by teacher prediction error and LIBERO average success.
    The Bottleneck-Projected Alignment Update obtains the lowest pseudo-action error and the best success rate, indicating that filtering student feedback is more effective than standard EMA.
    Middle: ablation of the VLA-specific reliability controller.
    Combining vision-language alignment, action feasibility, and temporal transition consistency produces the best performance while applying stricter pseudo-action filtering.
    Right: efficiency-performance trade-off.
    SemiVLA improves success rate within a small trainable-parameter budget, and SemiVLA with Selective LoRA achieves the strongest trade-off between adaptation cost and task performance.
    }
    \label{fig:reliability_teacher_efficiency}
\end{figure*}

Figure~\ref{fig:reliability_teacher_efficiency} further analyzes why SemiVLA improves over confidence-only self-training.
The full reliability controller selects fewer but more useful pseudo-actions, while the Bottleneck-Projected Alignment Update reduces teacher prediction error and improves the final policy.

\textbf{Component ablation.}
Table~\ref{tab:ablation_components} analyzes the contribution of each major component in SemiVLA. 
The supervised LoRA baseline follows the common parameter-efficient VLA adaptation setting used in recent VLA fine-tuning studies~\cite{kim2024openvla,kim2025fine,mitra2025mechanistic}. 
It obtains 81.0\%. 
Adding naive pseudo-action learning improves performance to 83.2\%, showing that unlabeled trajectories are useful even without sophisticated filtering. 
This is consistent with the general observation in semi-supervised learning that pseudo-labeling and consistency regularization can exploit unlabeled data~\cite{xie2020unsupervised,sohn2020fixmatch,wang2022freematch}. 
However, the gain is limited because noisy pseudo-actions can mislead the student.

When the VLA reliability controller is added, the average success rate increases to 86.1\%. 
This 2.9-point improvement over naive pseudo-action learning shows that pseudo-action quality is a key factor in semi-supervised VLA adaptation. 
The reliability controller filters pseudo-actions by vision-language alignment, action feasibility, and temporal consistency. 
This makes the unlabeled supervision more suitable for embodied control than confidence-only pseudo-labeling, which is commonly used in conventional SSL methods~\cite{sohn2020fixmatch,wang2022freematch}.

Adding the bottleneck-projected update further improves performance to 87.6\%. 
This shows that teacher evolution is also important. 
A teacher updated with noisy student feedback can become unstable, while the proposed update transfers only alignment-supported student directions. 
This design is related to teacher-student consistency learning~\cite{tarvainen2017mean} and is further motivated by bottleneck-style representation filtering~\cite{tishby2000information,DBLP:conf/iclr/AlemiFD017}. 
The full SemiVLA model obtains 87.9\%, which is 6.9 points higher than supervised LoRA. 
Therefore, Table~\ref{tab:ablation_components} confirms that self-distillation, VLA-specific reliability, and bottleneck-projected teacher update all contribute to the final performance.

\textbf{Reliability controller ablation.}
Table~\ref{tab:ablation_reliability} studies different reliability designs for pseudo-action selection. 
Using confidence alone gives an average success rate of 83.2\% and accepts 52.8\% of pseudo-actions. 
Although this accepts many unlabeled samples, the pseudo-actions are not always useful because confidence does not guarantee physical feasibility or visual grounding. 
This explains why confidence-only selection, despite being effective in classification-oriented SSL~\cite{sohn2020fixmatch,wang2022freematch}, gives only moderate improvement in VLA adaptation.

Single reliability signals improve performance slightly. 
Vision-language alignment alone reaches 84.0\%, action feasibility alone reaches 84.5\%, and temporal consistency alone reaches 84.8\%. 
These results show that each signal captures a useful aspect of pseudo-action quality. 
However, none of the single signals is sufficient, because VLA action prediction is jointly determined by language grounding, robot action constraints, and temporal dynamics~\cite{kim2024openvla,black2024pi_0,bjorck2025gr00t}.

Combining vision-language alignment and action feasibility improves the result to 86.0\%. 
Adding temporal consistency further improves the result to 87.9\%, which is the best performance. 
The acceptance ratio decreases from 52.8\% for confidence-only to 36.8\% for the full reliability design. 
This means the full controller is stricter. 
However, the final success rate is higher, showing that pseudo-action quality is more important than pseudo-action quantity. 
Table~\ref{tab:ablation_reliability} therefore supports the design choice of using multi-aspect VLA reliability instead of confidence-only pseudo-labeling.

\textbf{Teacher update strategy.}
Table~\ref{tab:ablation_update} evaluates different teacher update strategies. 
Keeping the teacher frozen achieves 85.2\% average success rate. 
This already performs better than supervised LoRA because the model can still benefit from reliable pseudo-action learning. 
However, a frozen teacher cannot evolve with the student, which limits the quality of later pseudo-actions.

Standard EMA improves the result to 85.9\%, but the gain is small. 
This suggests that simply averaging the student into the teacher is not enough for VLA adaptation. 
Although EMA teacher updates are widely used in teacher-student semi-supervised learning~\cite{tarvainen2017mean}, VLA adaptation requires more structured feedback because action prediction depends on semantic grounding and control precision. 
Reliability-aware EMA further improves the result to 86.8\% and reduces pseudo-action error from 0.171 to 0.154. 
This shows that controlling the update strength is helpful. 
However, it still uses a scalar update and cannot distinguish semantic alignment parameters from action precision parameters.

The proposed bottleneck-projected update achieves the best performance, reaching 87.9\% with the lowest pseudo-action error of 0.137. 
Its acceptance ratio is not the highest, but its pseudo-action error is the lowest and its final success rate is the best. 
This indicates that the update mechanism improves teacher quality rather than simply increasing the number of accepted pseudo-actions. 
By projecting student feedback through stable vision-language-action subspaces, the teacher becomes a more reliable pseudo-action generator. 
This is consistent with the general motivation of bottleneck-based learning, where nuisance-sensitive information is suppressed while task-relevant information is preserved~\cite{tishby2000information,DBLP:conf/iclr/AlemiFD017}. 
Thus, Table~\ref{tab:ablation_update} validates the need for a VLA-specific teacher update strategy.

\textbf{Generalization and robustness.}
Table~\ref{tab:generalization_robustness} evaluates whether SemiVLA improves generalization to unseen tasks and robustness to environmental variations. 
Robustness to distribution shift and perturbation has been widely studied in both visual recognition and robotic learning~\cite{hendrycks2019robustness,hendrycks2021many,tobin2017domain}. 
OpenVLA zero-shot obtains 49.6\% average success rate under these challenging settings. 
Supervised adaptation improves robustness, with OpenVLA + Selective LoRA reaching 66.7\%. 
This shows that task-specific fine-tuning helps the model adapt to the source tasks and related variations.

SemiVLA further improves generalization and robustness across all evaluation conditions. 
SemiVLA + LoRA reaches 74.8\%, outperforming OpenVLA + LoRA by 11.6 points. 
SemiVLA + Selective LoRA achieves the best average result of 77.8\%, which is 11.1 points higher than OpenVLA + Selective LoRA and 28.2 points higher than zero-shot OpenVLA. 
The improvement is consistent across unseen tasks, lighting variation, object variation, and distractors.

The gain on distractor robustness is particularly important. 
OpenVLA + Selective LoRA obtains 62.1\% under distractors, while SemiVLA + Selective LoRA reaches 73.9\%. 
This suggests that SemiVLA does not simply memorize source-task demonstrations. 
Instead, the reliability controller and bottleneck-projected update help the model learn more stable vision-language-action correspondences from unlabeled trajectories. 
This observation is aligned with prior robustness studies showing that stable representations and perturbation-aware learning can improve generalization under distribution shifts~\cite{zhou2022understanding,bai2021transformers,paul2022vision}. 
Table~\ref{tab:generalization_robustness} therefore shows that SemiVLA improves not only in-distribution adaptation but also robustness under realistic deployment shifts.
\section{Conclusions}
\label{sec:conclusion}

In this paper, we studied semi-supervised Vision-Language-Action (VLA) adaptation under limited supervision signals, where only a small subset of trajectories contains action labels and the remaining data provide action-unlabeled vision-language observations. 
To address this setting, we proposed SemiVLA, a self-distilled teacher-student framework that learns from reliable pseudo-actions on unlabeled trajectories. 
SemiVLA introduces a VLA-specific reliability controller to evaluate vision-language alignment, action feasibility, and temporal transition consistency, and further uses a Bottleneck-Projected Alignment Update to prevent noisy student feedback from contaminating the teacher. 
Experiments with OpenVLA and multiple adaptation strategies show that SemiVLA consistently improves over supervised fine-tuning and conventional pseudo-action self-training under limited action-label settings. 
These results demonstrate the value of reliable semi-supervised learning for data-efficient VLA adaptation.


\bibliographystyle{plainnat}
\setlength{\bibhang}{0pt}
\setlength\bibindent{0pt}
\bibliography{main}

\clearpage

\clearpage
\appendix
\startcontents[chapters]
\setcounter{page}{1}




\end{document}